\documentclass[12pt]{article}
\usepackage{graphicx, amsmath}
\usepackage{mathtools}
\usepackage{longtable}
\usepackage{booktabs}
\usepackage{multirow} 
\usepackage{mathrsfs}

\title{
	\textbf{Verbs On Action:} \\
	An Exploration on Multimodal Learning\\
}
\author{
        Student: Evin Pinar Ornek \\
        Advisor: Prof. Marie-Francine Moens \\
        Supervisor: Prof. Georg Groh \\
        KU Leuven \& TU Munich
}
\date{February 2019}

\begin{document}
\maketitle



\section{Introduction}

Humans learn language through several perceptive cues and an endless continuum of multimodal interactions. To learn the names of the objects around us, we need some kind of a supervision or a context. Either our parents explicitly point us the tangible, non-abstract objects in our first years, or we grab the meaning of the words from the peripheral context. Likewise, the movements of the objects are described by "verbs". We learn the meaning of the verbs by watching the objects in motion, or we grab a verb through a linguistic context without visually perceiving it. Then we use the learned objects and verbs in different unseen combinations, constitute novel sentences and generalize the verbs and nouns to new unseen instances or cases. There is an ongoing process of connecting, updating and renewing the inputs from different modalities \cite{barsalou18}.

In this work, we explore the possibilities of learning the \textbf{verbs} from multimodal cues in a similar way to humans and propose a neural network model that aims to jointly capture the visual and textual representation. The problem is to build a cross-modal joint space which will help retrieving a textual modal given a visual modal, or vice versa. If it is possible to create such a joint space that connects different modalities and types of inputs, it will be helpful in many arising areas such as video retrieval through natural language, event detection, video captioning, text generation from video, visual hallucination or synthesis from language... Furthermore, a human-like behavior on learning and capturing the multi-modal inputs can be defined as a weak machine learning problem such as unsupervised learning where there are no ground-truth labels, few-shot learning where the number of training samples is limited, or zero-shot learning where the test set includes unseen labels. Here, we especially focus on the zero-shot learning problem, and try to figure out if a joint multimodal space would help us in this problem. Such a space might be useful in generalizing over the unseen cases. 

Specifically, we focus on the action describing verbs and motions. On the visual side, since the action verbs have temporally rich features that differs for each action, we use the video inputs that reflect the spatio-temporal features of each action. On the language side, the actions are described by "verbs" and the state-of-the-art distributional language models reflect the semantic features of verbs through the co-occurence relationships with other words. Each action has a visual and a textual representation and we train two auto-encoders for these modalities. We train these auto-encoders with the paired visual-text samples together, so that they work in coordination and the latent vector between the encoders become the joint multimodal space. The auto-encoders are under complete which makes the latent vector a bottleneck that contains the main underlying structures of the data. The encoded text and encoded video should be the same or as close as possible, and the cross retrievals from text to video and vice versa should be applicable. 

It is shown that the current state-of-the-art zero-shot recognition is achieved through learnable linear mapping functions with a selection of different losses \cite{xianCVPR17}. We hypothesize that an auto-encoder will be able to better capture the modal representations than the compatibility mappers in \cite{xianCVPR17} because it will learn to reconstruct the modalities both separately and in a cross pass fashion. Since it will learn and reflect the structural differences of the modalities and their primary features, it will have a better semantic understanding and the zero-shot recognition will be possible. We introduce a minimalist model used with the atomic cues from different modalities to provide explainability of the structures and to explore the limitations and the requirements of such a zero-shot learner. Moreover, the two-way usage might add an additional practicality which could be used in different language-vision tasks.

The related work on visual action recognition, zero-shot activity recognition and the video event detection tasks are pointed out in section 2. The proposed model is explained in section 3 and the experiments with the results are given in section 4. The conclusion and the results are discussed in section 5.


\section{Background}

\subsection{Zero-shot activity recognition}
The zero-shot object classification has been applied through cross modal information transfer between the images and the language. For example, \cite{Socher13} projected the images onto semantic spaces, namely the word embeddings and \cite{Silberer14learninggrounded} used stacked autoencoders to ground the attributional features of objects to visual inputs. However, the research in zero-shot activity recognition is a relatively recent and less studied task.  

Zellers et al. tried classification of verbs in images, instead of objects, through given linguistic cues and hand-defined attributes \cite{Zellers17}. They proved to be able to predict the unseen actions from the images up to 42.17 top-5 accuracy. Youtube2Text mined the S/V/O triplets from the captions of videos, and try to predict the triplets from the videos in a hierarchical manner by minimizing the syntactic tree distances \cite{Guadarrama13}. 

Xu et al. performed unseen human activity classification with the help of videos. They use HOG and SIFT features along with the language in form of word embeddings to create a manifold through a transductive approach \cite{Xu15, Xu17}. However, there is an access to test labels during training in their setting. They test their approach on human activity datasets.

Piergiovanni et al. uses the sequential frames and sentences and encode them to a common representational space through temporal attentional layers they have defined in their previous work \cite{Piergiovanni18}. Our work is most similar with \cite{Piergiovanni18}, however instead of temporal attention module, we propose a simpler model to test the effectiveness of the auto-encoder neural network only which was not done on previous research. Then we examine the effects of different loss functions. \cite{Piergiovanni18} use sentences that describe the events, whereas we focus on the verbs and only use the relevant word vectors. Since we focus on atomic inputs instead of long sentences, we opt out the temporal attention. By starting from the small inputs and minimum sets of cues, we induce our problem as a distributional mapping problem to explore the cross-relations between different modalities. Our inquiry is on finding a mapping that will enable a generalization between the modalities to allow zero-shot class prediction.


\subsection{Action recognition}

The state-of-the-art video recognition approaches make use of 3D convolutional networks over sequential frames of either RGB format, or the optical-flow outputs. The recent works on are generally evaluated on common human activity datasets.  

Carreira et  al. introduced the Two-Stream Inflated 3D ConvNet (I3D) which combines the 3D frame features and 3D optical features for activity classification \cite{Carreira17}. Their model can be seen as an extended version of ImageNet that can recognize the spatio-temporal features through 3D streams. They test their model on Kinetics dataset and perform action recognition accuracies of up to $80.9\%$ on HMDB-51 dataset and $98.0\%$ on UCF-101 challenge \cite{kinetics}. 

Temporal Segmentation Networks try to predict the important segments over a video to predict the action \cite{Wang16}. They make use of the arrow of time in order to capture the differences between actions and classify them. Kong provides an in-depth literature review and summarization of different approaches along with their evaluations \cite{Kong18}. 

These models try to capture the visual semantics and classify the actions with the help of temporal dependencies and cues. However, these settings are not trained for zero-shot problem, and cannot find an unseen action class. Different than those, our work aims to be helpful in action classification task with the help of existing video recognition models, and examine the possibilities of grounding the language to the video. 


\subsection{Video event detection and captioning}

Gao et al. try to retrieve the specific events from the video by a language query \cite{Gao17}. The query comes in the form of a sentence, and the mapping is attained through Long Short Term Memory modules. Likewise, Hendricks et al. try to find the sequences events with the help of extra temporal words such as "after" or "during" that are used in the queries \cite{Hendricks18}.

The solutions to these tasks are specified according to how they define the problem and are hardly generalizable. However, they give rise to further questions on the relationship between video and language. Moreover, they share similarities with our work in terms of trying to understand the atomic particles of video by recognition models and align them with linguistic cues. Setting a multimodal joint space would be helpful in detecting the particular events from the video through the motionary words (verbs) and figuring out the sequential dependencies as well.

The dense video captioning problem focuses on atomic spatio-temporal features of videos to generate meaningful textual explanations . \cite{Li18, XuH18, Krishna17} make use of video recognition models along with language models and attention modules. However, (a) they require gold-datasets (perfectly explanatory captions for the video samples), (b) they hope the alignment of the particular words and parts of speech with the pixels through the help of attention modules, (c) they are not optimized for zero and few shot settings. In this case, our joint space might lead to an automatic alignment of some verbs to the temporal instances in videos, and become a starting point for alignments when there are no gold-datasets. This can be done by sliding the auto-encoder through the video. If the resulting vector has a high similarity to any verb, the latent vector can be used as a starter for an LSTM description generation model.


\section{Method} \label{problem}
In this work, we aim to incorporate the temporally rich visual cues and the relevant linguistic cues into a fixed sized embedding space. The videos provide the temporal information for the action verbs better than the static images, and a video understanding module can be used to extract these visio-temporal features. For the linguistic cues, the verbs from the distributional language models are used since they represent the words atomically in a large set of dimensions and capture the features of verbs on a similarity basis. We first describe how these cues are extracted, then explain the joint multimodal representation architecture.

\subsection {Video Understanding}

\begin{figure}[h]
\centerline{\includegraphics[width=1.1\textwidth]{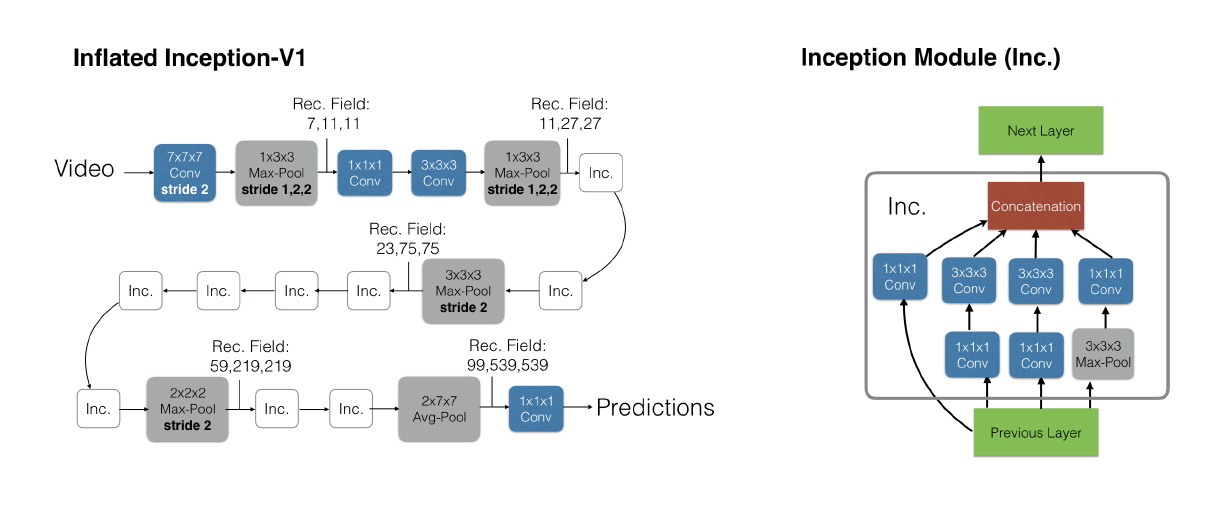}}
\caption{Inflated3D Action classification architecture by Carreira et al. \cite{Carreira17}, where Inc. labeled modules represent the Inception Module. On the last average pooling layer, there are 1024 features for t frames. t is smaller than the original size of T due to multiple convolutional layers and pooling operations. The predictions are calculated by pooling the last t frames and mapping to C classes. }
\label{i3d}
\end{figure}

Temporal dependencies can be captured best through sequences of frames (videos). For that, the state-of-the-art action classification networks make use of 3D convolutional networks on RGB features and optical flows and combine the predictions from both signals in a weighted manner. Here, we can use only one type of feature since our task is not about improving the visual classification. The aim of our study is to understand whether using such a network will be meaningful for our multimodal space. Figure \ref{i3d} demonstrates the network architecture. 

On the penultimate layer of the network, the original number of time steps $T$ is reduced to a much smaller size $t$, where each frame has 1024 features. On the last layer, these $t$ frames are averaged to a single vector and the classification is applied through softmax. Hence, we have a feature vector of size 1024 which represents the visual activity through the video. The original action recognition model uses these features to classify over the possible activity classes. Therefore, it means that this single vector is capable to reflect visio-temporal differences between actions. In this work, we use this vector as an input of the visual modality because it consists of rich information about the visual modality on a high dimensional setting.

\subsection {Textual Representation}

The distributional word embeddings have proven to be successful for the natural language processing tasks. Such embeddings include Glove or Word2Vec, which have similar distributional logic but trained in separate ways. In this work, we will use pre-trained Glove embeddings. There are different sets of embeddings trained on different datasets, and with different purposes. Each of them represent the co-occurence relationship of the words in the trained in-domain dataset. We choose the most common and generalizable embeddings with the highest dimensions. Besides these word embeddings, the state-of-the-art is shifting towards attentive models such as Elmo or Transformer \cite{bert}. In this project, we will only make use of a simple representative model because, again, our question is whether our multimodal space is meaningful or not. Yet, the better word models might improve the results further. 

\subsection {Joint Embedding Space}

We have two sources for multimodal understanding, (a) a video vector with \textit{C} features that represent the most important spatio-temporal features of the activity, (b) a word vector with \textit{D} dimensional embeddings. The video vector is extracted by passing the sequential frames to the 3D CNN layers of the action recognition network, whereas the word embeddings are learned through the skip-gram network. There are several approaches to find the relationship between them. 

One method is to find a direct mapping from video input to a word embedding. There, each video will be reduced to the number of dimensions of the word embeddings through either a fixed linear map or a neural network. However, as Collell et al. showed in their work, such mappings conserve the semantics of the input vectors rather than learning the common features along with the paired targets or mappings \cite{guillem2018}. In other words, they work biased towards the input vectors and will not be able to capture a symmetric relationship between different modalities.

\begin{figure}[h]
    \centering
      \includegraphics[width=0.95\textwidth]{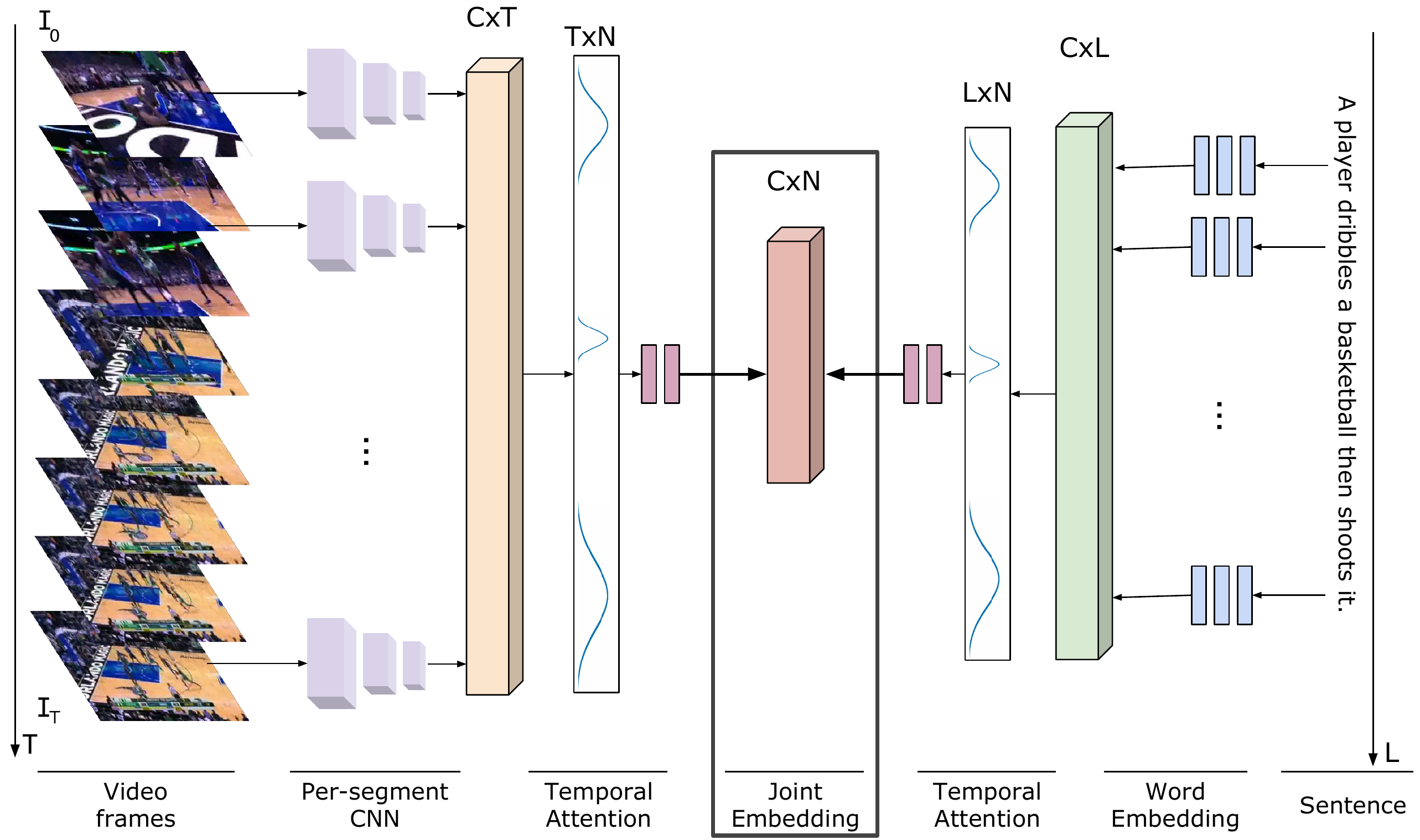}
      \caption{The multimodal embedding model used in \cite{Piergiovanni18}. The video is of size \textit{CxT} where \textit{C} is the number of features and \textit{T} is the number of time steps. Likewise, the caption sentences are embedded by Glove vectors and their size is \textit{CxL} where \textit{C} is the number of Glove features and \textit{L} is the number of words. In the encoder/decoder layer, they use the temporal attention mechanism to reduce the time dimension \textit{T} to \textit{N} and there are feed forward layers after the encoders. }
      \label{fig:model}
\end{figure}

An approach to overcome the inherent limitations of the direct mapping between multimodal vectors might be to build up an embedding space in between the videos and words and incorporate a sophisticated loss function to capture the model. A neural network in this direction is proposed by Pergiovanni et al. is illustrated in the Fig. \ref{fig:model}. 

They introduced the autoencoders to create an embedding space which are described as: 
\begin{align*}
\mbox{\textbf{Video Encoder} } & E_{V}: v \mapsto z_v
& 
\mbox{\textbf{Video Decoder} } & G_{V}: z \mapsto v
\\
\mbox{\textbf{Text Encoder} } & E_{T}: t \mapsto z_t
&
\mbox{\textbf{Text Decoder} } & G_{T}: z \mapsto t
\end{align*}

The \textit{"G"} stands for "Generator", whereas \textit{"E"} for "Encoder". Their model is trained with a mixture of these loss functions:

\begin{equation}
\label{eq:recons}
    \mathcal{L}_{recons}(v,t) = ||G_V(E_V(v)) - v||_2 + ||G_T(E_T(t)) - t||_2
\end{equation}

\begin{equation}
\label{eq:joint}
    \mathcal{L}_{joint}(v,t) = ||E_V(v) - E_T(t)||_2
\end{equation}

\begin{equation}
\label{eq:cross}
    \mathcal{L}_{cross}(v,t) = ||G_T(E_V(v)) - t||_2 + ||G_V(E_T(t)) - v||_2
\end{equation}


\begin{equation}
\label{eq:paired}
    \mathcal{L}(v,t) =  \alpha_1\mathcal{L}_{recons}(v,t) + \alpha_2\mathcal{L}_{joint}(v,t) + \alpha_3\mathcal{L}_{cross}(v,t) 
\end{equation} 

Here, the video encoder learns to construct it's own input, and the text encoder learns to construct the textual input through the $L_{recons}$ loss.  Whereas $L_{joint}$ enforces the constructed embedding space from different modalities to be as close as possible. The aim is to match the representative vectors coming from both modalities. In addition, the $L_{cross}$ connects the video encoder to the text decoder, and the text encoder to the video decoder since we wish a text-visual sample to be retrieved from each other. This loss is commonly used in computer vision tasks to transfer the artistic style of an image to another image. 

Different from \cite{Piergiovanni18}, in order to check the limits and constraints of such a two-way auto-encoder model for two income modalities, we introduce a set of feed forward layers and nonlinearities instead of the attention. In our model, the neural layers constitute a bottleneck layer between the two modalities which represent the joint semantic space. Moreover, in \cite{Piergiovanni18}, the attention module summarizes the video features over the time dimension, and likewise, summarizes the word features from a paragraph. In this work, we use atomic inputs so that there is a 1-dimensional vector representing the video features instead of a matrix, and there is a word vector for the corresponding action word instead of a caption or a paragraph. Hence, there are no temporal attention networks, because we assume that the temporal features of an action is already included in its feature vector. 

Our auto-encoder is trained and tested with the loss functions described above ($L_{recons}$, $L_{joint}$ and $L_{cross}$). In addition to these losses, the model will be able to learn better with the negative samples. We wish a non-related class vector to be far from the true class vector in the shared space. The authors of  \cite{Piergiovanni18} have introduced the negative learning through additional discriminators for each modal space, and the adversarial loss functions. In this work, we have chosen a simpler approach and used a margin ranking loss:

\begin{equation}
\begin{split}
\label{eq:rank}
    \mathcal{L}_{rank}(s_{1},s_{2}) = max( 0, margin - (s_{1} - s_{2})) 
\end{split}
\end{equation}

Here, $s_{1} = cos( E_V(v) , E_T(t))$ is the similarity between the constructed vector from a paired text and video. Whereas $s_{2} = cos(E_V(v_{n}) , E_T(t_{n})) $ is the similarity between the unpaired text and video. We wish the $s_{1}$ be higher than $s_{2}$. The margin ranking loss enforces this with a predefined margin. A similar function is used by Zellers et al. \cite{Zellers17} for the textual inputs only.


\section{Experiments}

\subsection{Datasets}

The Kinetics Human Action Video Dataset consists of 400 different human activity classes with 10 seconds of videos \cite{kinetics}. These activities cover a broad range of movements, such as sports(playing basketball, snowboarding), basic body motions(jumping, clapping), eating or cooking related activities, hobbies, communicative motions etc... We used this dataset as video input and extracted the feature vectors through the official pre-trained I3D model released at \footnote{ https://github.com/deepmind/kinetics-i3d/}. Each extracted video has 1024 features. 

As a textual input, the Glove word embeddings of each activity class is used. Some activities include several words such as "playing piano". Here, we followed Iyyer et al's approach \cite{IyyerMBD15} and averaged the Glove vectors over these words. The pretrained Glove embeddings can be found.\footnote{https://nlp.stanford.edu/projects/glove/} The 300-dimensional vectors trained on Wikipedia 2014 are used. 

In addition to the video and word embeddings, we tested the model on the IAPR TC-12 dataset which consists of 200K images and their captions. The features of these images are extracted through VGG-128 image recognition network and collect the last layer of 128 dimensional feature space. The textual features are extracted by the help of bidirectional gated recurrent unit(bi-GRU) and has 64 dimensions. We have used the pre-trained feature vectors from \footnote{https://github.com/gcollell/neural\_cross\_modal\_maps}. Even though the main purpose of our model is to learn the temporal visual inputs, such static inputs will help to improve and test the model further. 

The IAPR TC-12 dataset is split to 16K train, 2K validation and 2K test. The Kinetics Dataset has 400 videos for each 400 action classes. In this work, a smaller subset is generated by randomly selecting 300 classes and 40 videos for each of the class. In total, there are 10K train, 1K validation and 1K test data.  

\subsection{Model and Implementation Details}

Each of the autoencoders consist of 3 feed-forward layers of decreasing sizes, with the ReLU non-linearity and drop-out between each FF layer. The joint space has a smaller number of dimensions: for the Kinetics dataset of 1024-d vision and 300-d text inputs. The bottleneck in between is tested with sizes of 300-d, 200-d and 150-d. Likewise, for the IAPR TC-12 dataset, the joint space size for 128-d vision and 64-d vision is tested with 64-d and 48-d. 

The autoencoders are trained with the Adam optimizer, started with a learning rate of $1e-3$ with a weight decay of $1e-5$. A higher starting learning rate is also tried, but the model ended up in a local optima. The drop-out rate of 0.5 is used, though there was no major effect of different rates. The best models during training are selected by the lowest validation error. 

\subsection{Evaluation and Results}

First, we measured the action class prediction accuracy in order to be able to understand whether our model was successful at representing the multi-modal information from two inputs. For that, we decoded the test video vectors into textual vectors, and retrieved the most similar N word vectors according to cosine similarity metric. If the actual class is in the set of most similar vectors, it is counted towards a hit, hence contributing for the \textbf{top-N accuracy}. In our case, the ground truth class might have several words ("playing basketball", "making sandwich", "walking with horse"...) and is counted as a hit if any one of the words is matched. It is important to note that the tests are not in the "Generalized Zero-Shot Learning(GZSL)" setting, and the similarity is applied in the global word embedding space, not restricted by only the classes. Therefore, our evaluation setting may not directly conform with the current trends on zero-shot evaluation.

Second, the zero-shot action recognition is measured to evaluate the generalization capability of the model over the unseen classes. In this case, the test set consists of randomly selected 10 unseen classes with 40 instances each. We did not include the seen class instances in the test set in order to evaluate the zero-shot accuracy explicitly. Again, we calculated the top-N accuracies.

In Table \ref{tab:results}, the prediction accuracy for seen and unseen classes are reported. The hyper-parameters of each loss function is adjusted to examine the effects of different combinations. Each model is trained for 300 epochs. The first five lines show the results for the test setting where there are 300 classes with 40 video instances. In addition, we have extended the tests with two more settings where the results are shown at the bottom two rows of Table \ref{tab:results}. The first setting included 100 training classes with 100 videos for each instance. The number of unseen classes is increased to 30. In the second setting, the number of training classes is increased to 200 with 100 instances each, and again with 30 unseen classes. 

\begin{table}[htbp]
\makebox[\textwidth][c]{
\begin{tabular}{ccc|ccc|cc}
\toprule
    &  &  & \multicolumn{3}{c}{Seen Classes} & \multicolumn{2}{|c}{Unseen Classes} \\ 	
    \cline{4-6}\cline{6-8} 
    $L_{recons}$ & $L_{rank}$ & $L_{cross}$ &      top-5 & top-10   &  top-30  &  top-5 & top-10  \\ \midrule
    1			&	0 	     & 	      0             &  12 & 20 & 30 & 0 & 12 \\
    1			&	1 	     & 	      0             &  7 & 10 & 18 & 10 & 14  \\
    0.1			&	1 	     & 	      0             &  3 & 4 & 9 & 5 & 7  \\
    1			&	1 	     & 	      1             &  6 & 8 & 21 & 8 & 9 \\
    0.1			&	1 	     & 	      1            &  6 & 10 & 21 & 8 & 9  \\
    \hline
    0.1			&	1 	     & 	      1            &  11 & 16 & 29 & 12 & 18  \\
    0.1			&	1 	     & 	      1            &  8 & 12 & 23 &  9 &   11 \\
    
\bottomrule
\end{tabular}
}
\caption{ Results on the action class prediction averaged over the test sets (in \%). The top-5, top-10 and top-30 accuracies are given for the seen class prediction on the test dataset, whereas the top-5 and top-10 accuracies are given for the unseen class prediction task.}
\label{tab:results}
\end{table}

As it can be seen from the table, the model with only the reconstruction loss gave the highest top-N accuracies on seen action prediction task. Then, the addition of ranking loss only did not improve the results, and higher focus on this loss has decreased the retrieval. The combination of all losses, with a higher influence of cross and ranking losses worked best among the multi-loss models. However, in the zero-shot prediction case, the equally weighted combination of $L_{recons}$ and $L_{rank}$ showed the highest top-5 and top-10 accuracy.

In the first additional test where there are less classes but more data per class, the accuracies for seen class prediction were as high as the initial experiments, and the zero-shot results both for top-5 and top-10 were the best. In the second setting where there are more classes with more data, however, the model did not perform better than in the initial experiments. Probably, it learned a representation with a fixed set of weights that is optimal for a highest score (local optima), but does not vary much over different inputs. This would mean that it did not have a generalization capability, and could not be fixed with the validation split nor drop-out. 

One of the unexpected observations was the effect of $L_{joint}$ loss. During training with this loss, the model stopped learning after several iterations. Moreover, it gave a single fixed vector output for any test sample without depending on the input class. Hence, we believe that such a loss used with the combination of others finds a fixed solution and loses the ability to generalize and vary on different inputs. 

Here, since there is no baseline approach, it is not possible to compare these results with another model. However, a fully linguistic attribute based zero-shot model of \cite{Zellers17} reported the top-1 accuracy as $18.15$ and top-5 as $40.17$. They extract the verb's linguistic properties such as being a motion, having social aspect, the expected duration, and map them to static image features. Our results might indicate that such a simple auto-encoder model is not as successful to predict the classes of unseen actions based only on the visual inputs and the word embeddings.


\subsubsection*{Nearest Neighbor Overlap}

Furthermore, in order to compare the effect of each input modality on the representation space, we have used the mean nearest neighbor overlap measure (mNNO) \cite{guillem2018}. The mNNO is defined as:

\begin{equation}
	\text{\textit{\textbf{mNNO}}}^K(V,Z) = \frac{1}{KN} \sum_{i=1}^N \text{NNO}^K(v_i,z_i)
\end{equation}

where $V = \{v_i\}^{N}_{i=1}$ and $Z = \{z_i\}^{N}_{i=1}$ are two sets of $N$ paired vectors. The $\text{\textbf{NNO}}^K(v_i,z_i)$ indicates the number of common vectors in the K nearest neighborhoods of $v_i$ and $z_i$. For instance, let the nearest 3 neighbors of $v_{cat}$ be $\{v_{dog}, v_{tiger}, v_{lion} \}$ and $z_{cat}$ be $\{z_{mouse}, z_{tiger}, z_{lion} \}$. The intersection of their neigborhood is $\{tiger, lion\}$ for K=3, and the mNNO score is $2/3$.

In our setting, when we decode a textual vector from a visual vector, the number of neighbor overlaps for these two vectors, and the neighbor overlaps between the ground truth word vector and the decoded word vector should be ideally close to each other. This will mean that the model can learn equal amount of information from different modalities instead of reflecting the topology of only one or the other.

We compared the mNNO scores from text-to-video and from video-to-text, with both the auto-encoder and the linear mapping settings. We have first measured the scores for IAPR TC-12 dataset for clear explanations. Then, we calculated for the Kinetics actions dataset. The results are reported in Table \ref{tab:results2}.

\begin{table}[h]
\makebox[\textwidth][c]{
\begin{tabular}{@{}lccccc@{}}
\toprule
	&  &  & $X,f(X)$ & $Y,f(X)$ \\ \midrule
	\multirow{4}{*}{\rotatebox[origin=c]{90}{IAPR TC}} & \multirow{2}{*}{$V\rightarrow T$} & \textit{ff} & \textbf{0.428}$$ & 0.162  \\
	&  & \textit{AE} & {0.097}$$ & \textbf{0.257}  \\ \cmidrule(l){2-5} 
	& \multirow{2}{*}{$T\rightarrow V$} & \textit{ff} & {0.049}$$ & 0.049  \\
	&  & \textit{AE} & {0.1}$$ & \textbf{0.236}  \\ \midrule
	
	\multirow{4}{*}{\rotatebox[origin=c]{90}{Kinetics}} & \multirow{2}{*}{$V\rightarrow T$} & \textit{ff} & \textbf{0.349}$$ & 0.109  \\
	&  & \textit{AE} & \textbf{0.144}$$ & 0.079  \\ \cmidrule(l){2-5} 
	& \multirow{2}{*}{$T\rightarrow V$} & \textit{ff} & {0.056}$$ & \textbf{0.059}  \\
	&  & \textit{AE} & {0.081}$$ & \textbf{0.149} \\ 
	
\bottomrule
\end{tabular}
}

\caption{Mean nearest neighbor overlap scores for video-to-text and text-to-video transfers for the test dataset. \textit{ff} indicates the neural feed forward layer, whereas AE is our two-way auto-encoder. \textit{X} represents the input, \textit{f(X)} the mapped output, and \textit{Y} the ground truth. }
\label{tab:results2}
\end{table}

It can be seen that for the IAPR TC-12 dataset, the auto-encoder learns the modalities similar to the output modality without depending whether it is a video or a text, though the neural mapper sustains the features of the input modality. For the Kinetics dataset, no matter if it is a linear mapping or an auto-encoder, the transferred features are more similar to video modality. The reason behind this might be the high difference in the number of visual and textual dimensions (1024 versus 300).

\subsection{Discussion}

The attempt of a joint space which would make the transfer of modalities in each directions without the requirement of extra manually defined attributes might be useful in video and language related tasks. However, our experiments on the prediction tasks showed that such a model built in our described approach was not capable of retrieving high results. 

\subsubsection*{Model limitations}

We used two stacked auto-encoders consisting of 3 neural layers each with ReLU for nonlinear learning. Each auto-encoder learns the latent features of the inputs while learning to reconstruct them back. The latent feature space acts as a semantic space that learn the shared distributional features among the modalities. The effects of model features:

\begin{enumerate}
\item The joint loss forced the networks to have a fixed embedding without generalizability. 
\item  The cross loss taught the model to retrieve the related item across the encoders. The effect of this has been tested in the preliminary experiments. Without the cross loss, the model cannot bridge the modalities.
\item  The ranking loss better enhanced the ability to differ between unrelated items and added an extra cue for "sharedness".
\item The drop-out made the model better generalizable and relieved the effect of \textit{hubness} to some extent. Without drop-out, the classification results did not have a large variety and pointed to same set of words. 
\item Non-linearity increased the accuracy on the preliminary experiments, which might be related with increasing the learnability of the model.   
\end{enumerate}

\indent The existing zero-shot \textbf{object} classification problem is shown to have higher accuracy with compatibility learning models that learn the mapping between the distributions rather than the attribute classifiers \cite{xianCVPR17}. However, zero-shot \textbf{activity} detection models' state-of-the-art accuracies are achieved by the help of manually defined linguistic attributes of verbs \cite{Zellers17}. The compatibility learning models map the different modalities to each other through the help of joint loss and a ranking loss. In this work, we hypothesized that a two-way cross mapper would improve the zero-shot accuracy, include equal level of information from both modals and provide a holistic solution to other cross-modal tasks.

\subsubsection*{Zero-shot recognition}

The question of whether it is possible to map the classification based visual distributional models to the neighborhood based textual distributional models without any extra prior knowledge is still remaining. In other words, we casted the problem as follows:

\textit{ \textbf{Visual space}: Each action's visual features are represented over V dimensions and could be clustered according to action classes. \\
\indent \textbf{Textual space}: Each word can be represented by D features where the similar words will be closer to each other in the space. \\
\indent \textbf{Zero-shot mapper}: A learnable mapper from the visual space to the textual space which could be generalized for unseen classes of visual space. }

Here, the main difference between the spaces is that, the visual space does not consider the similarity across different actions, but it only contains the class-wise differences which is helpful for classification. Whereas the textual space has a rich neighborhood that shows the similarity between each word. Hence, the textual modality should provide a cue for similarity relationship to the multi-modal representation space. The visual classes will then gain semantic information and the zero-shot recognition will be possible(or vice-versa). 

Another difference between the spaces is the types of information they include. The textual embeddings include any types of words, not only verbs but also nouns, adjectives, pronouns, adverbs... The visual space, on the other hand, contains information on geometric shapes, edges, textures and depth. Would it be really meaningful to try to map these different types of informations to each other? Can the zero-shot classification problem solved by such a mapper? This question is directed to both our work, and to the current research direction on zero-shot learning.

\subsubsection*{Possible improvements} There might be extra options to improve our model: 

\begin{itemize}
  \item It can be extended with a discriminative loss where there is an additional model that learns to separate the real versus fake visual input and the encoder competes in order to trick the discriminator. This way, the encoder will learn the underlying structure of the visual data better and will have a higher capacity to predict either seen or unseen classes.
  \item The number of words in the textual space can be restricted to only verb vectors for practicality.
  \item A different language model, either trained on a more related corpus that especially consists of more sports or activity related words might improve the accuracies.
  \item The current action recognition model makes use of all of the visual input frames where the background and other objects are included. A better approach would be to have segmented frames or frames with bounded boxes through an object tracker. At the moment, such a dataset does not exist for a large set of activity classes. 
  \item The neural networks may not be helpful to solve the problem, hence, different probabilistic cues could be introduced. 
\end{itemize}





\section{Conclusion}

In this work, we have proposed an auto-encoder based neural model that aimed to connect the multimodal representations over a joint space. We focused on short activities, their video representations and the distributional word features. Such a joint space which would make the transfer of modalities in each directions useful in video and language related tasks. The action class prediction experiments showed that our model was not capable of achieving results high enough to successfully assist on different tasks. There might be possible points to improve this model such as extending with discriminative loss, using a different language model, or a video recognition model. However, overall, our model used an approach which did not exist in the literature of zero-shot learning and activity classification, and the accuracies indicate that the research in this direction might be promising.  


\bibliographystyle{abbrv}
\bibliography{mybibfile}

\end{document}